\documentclass[final]{cvpr}

\usepackage{times}
\usepackage{epsfig}
\usepackage{graphicx}
\usepackage{amsmath}
\usepackage{amssymb}
\usepackage{multirow}
\usepackage{bm}

\usepackage{float}

\usepackage{CJK}
\usepackage[top=2cm, bottom=2cm, left=2cm, right=2cm]{geometry}
\usepackage{algorithm}
\usepackage{algorithmicx}
\usepackage{algpseudocode}
\usepackage{amsmath}
\usepackage{indentfirst}
\usepackage{array}
\newcommand{\PreserveBackslash}[1]{\let\temp=\\#1\let\\=\temp}
\newcolumntype{C}[1]{>{\PreserveBackslash\centering}p{#1}}
\newcolumntype{R}[1]{>{\PreserveBackslash\raggedleft}p{#1}}
\newcolumntype{L}[1]{>{\PreserveBackslash\raggedright}p{#1}}

\usepackage[pagebackref=true,breaklinks=true,colorlinks,bookmarks=false]{hyperref}

\def\cvprPaperID{3962} % *** Enter the CVPR Paper ID here
\def\confYear{CVPR 2021}
\usepackage{lipsum}

\makeatletter\renewcommand\paragraph{\@startsection{paragraph}{4}{\z@}
  {.5em \@plus1ex \@minus.2ex}{-.5em}{\normalfont\normalsize\bfseries}}\makeatother

\newcolumntype{x}[1]{>{\centering\arraybackslash}p{#1pt}}
\newlength\savewidth\newcommand\shline{\noalign{\global\savewidth\arrayrulewidth\global\arrayrulewidth 1pt}\hline\noalign{\global\arrayrulewidth\savewidth}}

\begin{document}

%%%%%%%%% TITLE
\title{Unsupervised Part Discovery via Feature Alignment}

\author{Mengqi Guo\textsuperscript{1}, Yutong Bai\textsuperscript{2}, Zhishuai Zhang\textsuperscript{2}, Adam Kortylewski\textsuperscript{2}, Alan Yuille\textsuperscript{2}\\
\textsuperscript{1}Beihang University\quad\textsuperscript{2}Johns Hopkins University\\
{\tt\small\{im.guomengqi, ytongbai, zhshuai.zhang, akortyl1, alan.l.yuille\}@gmail.com}\\
}

\maketitle
\pagestyle{empty}
\thispagestyle{empty}

%%%%%%%%% ABSTRACT
\begin{abstract}
   Understanding objects in terms of their individual parts is important, because it enables a precise understanding of the objects' geometrical structure, and enhances object recognition when the object is seen in a novel pose or under partial occlusion. 
   However, the manual annotation of parts in large scale datasets is time consuming and expensive.
   In this paper, we aim at discovering object parts in an unsupervised manner, \ie, without ground-truth part or keypoint annotations. 
   Our approach builds on the intuition that objects of the same class in a similar pose should have their parts aligned at similar spatial locations.
   We exploit the property that neural network features are largely invariant to nuisance variables and the main remaining source of variations between images of the same object category is the object pose.
   Specifically, given a training image, we find a set of similar images that show instances of the same object category in the same pose, through an affine alignment of their corresponding feature maps. 
   The average of the aligned feature maps serves as pseudo ground-truth annotation for a supervised training of the deep network backbone.
   During inference, part detection is simple and fast, without any extra modules or overheads other than a feed-forward neural network. 
   Our experiments on several datasets from different domains verify the effectiveness of the proposed method. For example, we achieve 37.8 mAP on VehiclePart, which is at least 4.2 better than previous methods.
\end{abstract}

\section{Introduction}
Computer vision is to a large extent concerned with the analysis of objects in images.
An important challenge in object analysis is to learn representations that are robust to partial occlusion, or changes in the object pose and appearance.
Part detectors have emerged as an important intermediate object representation that facilitates to achieve robustness to these variations. 
As a result, part detectors have become of central importance for many visual understanding tasks, including viewpoint estimation \cite{pavlakos20176}, human pose estimation \cite{cao2017realtime}, action recognition \cite{messing2009activity}, feature matching \cite{long2014convnets}, image classification \cite{zhang2014part}, and 3D reconstruction \cite{kanazawa2018learning}. 
While part detectors are very useful for computer vision, a main limitation is that detectors are typically learned using large amounts of annotated training data \cite{DBLP:conf/cvpr/CaoSWS17, huang2020interpretable,wang2017detecting, DBLP:conf/cvpr/ZhangXWXY18}.
However, in practice, obtaining such annotations is very time consuming. Therefore, it is difficult to scale fully supervised approaches to novel categories, which raises the need for unsupervised learning approaches.

The main challenge for unsupervised approach to learn part detectors is that they need to learn a set of features/neurons corresponding to semantically meaningful parts tightly on a single image, and consistently across different images of the same object class.
However, objects in images are highly variable due to changes in their pose, shape, and texture, as well as due to changes in the scene illumination, and the background clutter. The same parts may have slightly different appearance at different images, and it's non difficult to learn that without supervision.

\begin{figure}\includegraphics[width=\linewidth]{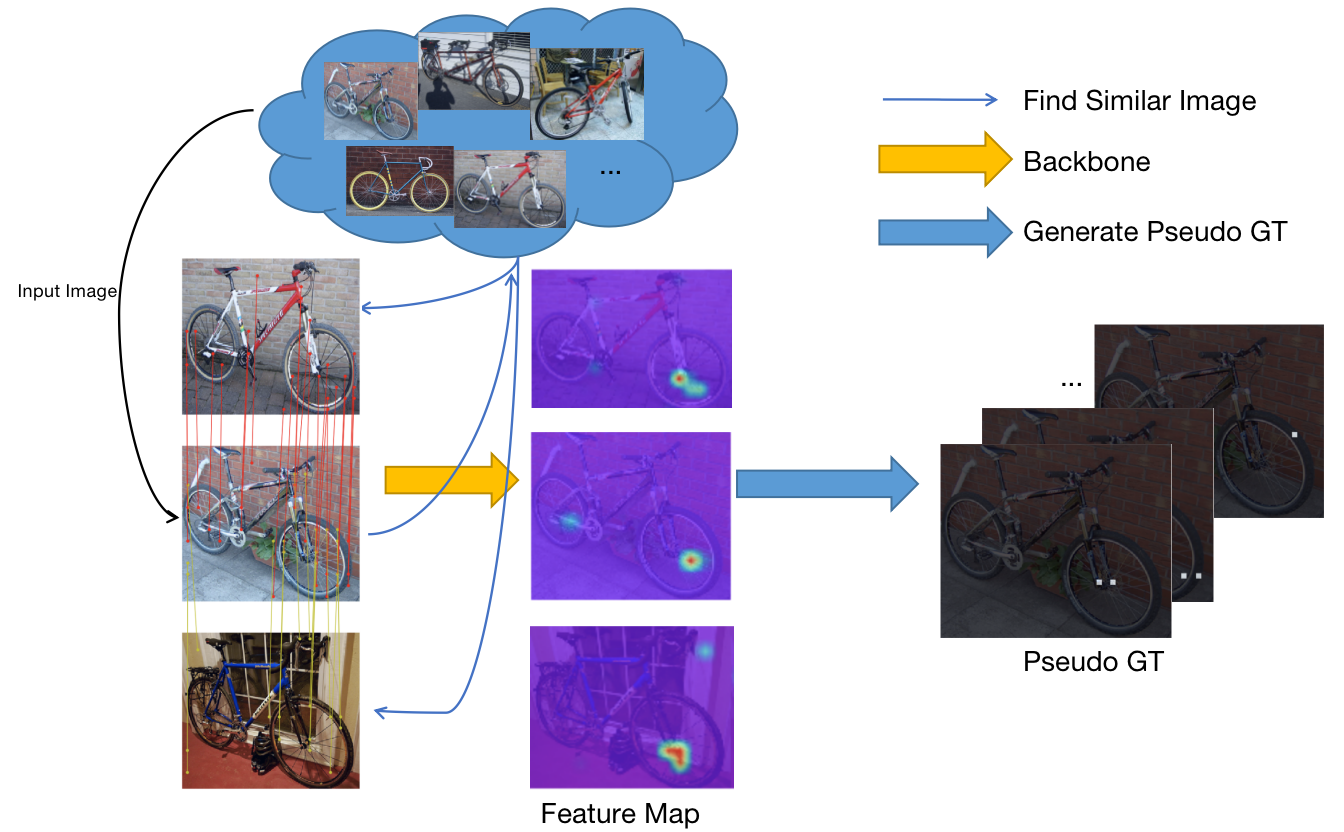}
\caption{Examples demonstrate the intuition of our method. Upper left: training image set. Lower left: a training image and two similar images, with distinguishable keypoints matched. Lower middle: the heatmap of an output channel, which corresponds to the front wheel, but with some false positives. Lower right: pseudo ground-truth generated from the training image and the similar images, for different output channels. This could be used to supervise the output feature maps shown in the lower middle.
}\label{fig:teaser}
\end{figure}

In this paper, we propose an unsupervised framework for learning part detectors.
Based on a collection of images of any object type, we learn part detectors that are consistent across different object instances, 3D poses and articulations.
The key intuition underlying our work is that part detectors should fire consistently across different images, w.r.t. some spatial transforms. This should be true especially for those images which are roughly similar with each other, as shown in Figure~\ref{fig:teaser}. This suggests that even without ground-truth part annotation, there still exists supervisions, coming from the coherence between similar images. To this end, we propose our training approach with three steps. 1) Given a training image, find a set of other training images similar to it, by comparing the CNN embedding of any off-the-shelf ImageNet-pretrained models. This allows us to enforce coherence of part detectors easier, without dealing with large transformation among images. 2) Align the similar images with the training image with some spatial transformation, by exploiting matched feature pairs (\eg, the features at wheels, window corners locations are more distinguishable to be matched). The strongly matched feature pairs can suggest the spatial correspondence between the training image and the similar images. 3) Finally, we can encourage coherence between the feature maps (those used for part detection) of the training image and the similar images. To achieve that, we generate a pseudo ground-truth map from the average of the aligned feature maps, and use that pseudo ground-truth map to supervise the feature map of each of the images. We also consider the intuition that the part detectors should fire sparsely on each image, so we add some constraints when generating the pseudo map by limiting the number of locations each feature map channel can correspond to. This can avoid a trivial solution, and allow the learned part detectors to be diverse and cover as more actual parts as possible.

Following the common scheme of related work on unsupervised part detection, we quantitatively evaluate our unsupervised part detection method for several different datasets on different domains, including VehiclePart~\cite{wang2017detecting}, CelebA~\cite{celeba}, CUB~\cite{CUB}, where we achieve state-of-the-art performance, as well as MAFL~\cite{MAFL}, where we perform comparably to related work.
Our qualitative results show that our model predicts part landmarks precisely and consistently even in different camera viewpoints and object poses.

\section{Related Work}
Recently, with the emergence and development of deep neural networks~\cite{lecun2015deep}, object recognition\cite{DBLP:conf/nips/DaiLHS16,DBLP:conf/cvpr/sangBS17, DBLP:conf/eccv/JiangLMXJ18, DBLP:conf/nips/RenHGS15} has achieved great success. As a detailed version of object recognition, part discovery and detection have been receiving more and more research attentions, for different domains, such as vehicles~\cite{bai2019semantic, wang2017detecting}, faces~\cite{zhang2014facial}, humans~\cite{newell2016stacked}. However, this is still a relatively less studied task, mainly due to the lack of annotations, which could be magnitudes of more expensive to obtain than object-level annotations.

Previous works for unsupervised part detection, including~\cite{wang2017detecting, wang2017visual, zhou2014object}, show that parts usually have similar local appearances, can be captured by individual neurons~\cite{zhou2014object} or k-means clusters~\cite{wang2017visual} of internal feature vectors. We choose clustering of internal feature vectors as our baseline, by adding one more layer to the backbone with the cluster center features as the layer weights. However, different parts and background may have similar local appearance, which could lead to fuzzy clusters. Other works find that parts can be supported by surrounding contexts, and can be detected by in voting scheme~\cite{wang2017detecting}. Some researches (\eg~\cite{thewlis2017unsupervised})
reformulate the landmark detection problem into image deformation. Nevertheless, those methods mainly take only the feature inside individual images or feature maps into consideration, while ignoring the similarity and consistency across different images or feature maps. Related work~\cite{hung2019scops} considers the local feature as well as semantic consistency between images. They build geometric concentration loss to concentrate the segmentation from fragments and equivalence loss which is based on the hypothesis that images have same part segmentation predictions with regard to the corresponding transformation. Then, they also build semantic consistency loss with a set of semantic part bases, which does not take spatial alignment into account simultaneously. On the other hand, our method exploits the spatial correspondence between similar images, from the intuition that similar images should have the same parts at similar locations.

Since unsupervised part detection is difficult, some works address the few-shot learning task. \cite{bai2019semantic} is one representative example. They use a few training images with ground-truth annotations and build 3D models to bridge the testing image to those training images with ground-truth. However, in addition to the requirement of labels, their model could be slow since they run a matching algorithm during the inference. We would like to build a simple and efficient model so that we may run the inference with a little overheads. Another work of weakly supervised part detection is \cite{huang2020interpretable}. They detect part landmarks with a simple prior of the occurrence of object parts which were trained with image-level labels.

\begin{figure*}[!ht]
  \includegraphics[width=\linewidth]{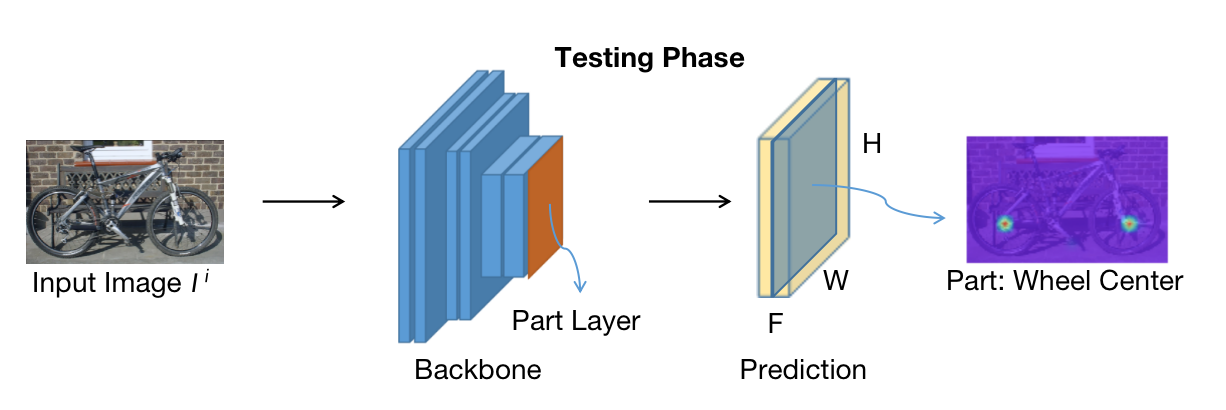}
  \caption{Our model is illustrated in the picture, the image is a 2D image with an object (bicycle as an example on the left), and the output feature map of the backbone is semantically meaningful and corresponds to object parts (bicycle wheel center as examples on the right).}
  \label{fig:test_phase}
\end{figure*}

\section{Proposed Method}
\label{Method}
The goal of our paper is to detect keypoints or parts of an object on a given 2D image, with a convolutional neural network (CNN). Without training on ground-truth part labels, we expect the output feature map of the CNN to be semantically meaningful and able to correspond to parts precisely. As shown by the example in Figure~\ref{fig:test_phase}, the input on the left is a 2D bicycle image, and the output feature maps on the right can capture object parts.

Formally, given an input image noted with $\bm{I}\in\mathbb{R}^{H\times W\times 3}$, we want to generate a semantically meaningful output feature map $\bm{O}\in\mathbb{R}^{h\times w\times c_o}$ to capture the parts of the corresponding object where $h$, $w$, and $c_o$ correspond to the height, weight, and number of channels of feature map ($h$ and $w$ can be then up-sampled to $H$ and $W$, the size of image).
The ground-truth part annotations of the image $\bm{I}$ with $n$ different parts are $\bm{P}=\{p_1,p_2,...,p_n\}$, and are used for evaluation only. 
$p_i=[(x_{i1}, y_{i1}), ...]$ corresponds the locations of the $i$-th part and may contains more than one item because there may be multiple instances of some parts (\eg wheels) and each item corresponds to the center of an instance of the part in the image. To detect the parts, we expect each of the part $p_i$, should be corresponded by at least one channel of $\bm{O}$, whose values are high around the part center $(x_{ij}, y_{ij})$ and are low for the areas far away from these part instance centers. The rightmost image in Figure~\ref{fig:test_phase} shows an example, where $p_1$ of the bicycle image is corresponded by the feature map channel illustrated in the heat-map.

\subsection{Overall Framework}

\begin{figure}\includegraphics[width=\linewidth]{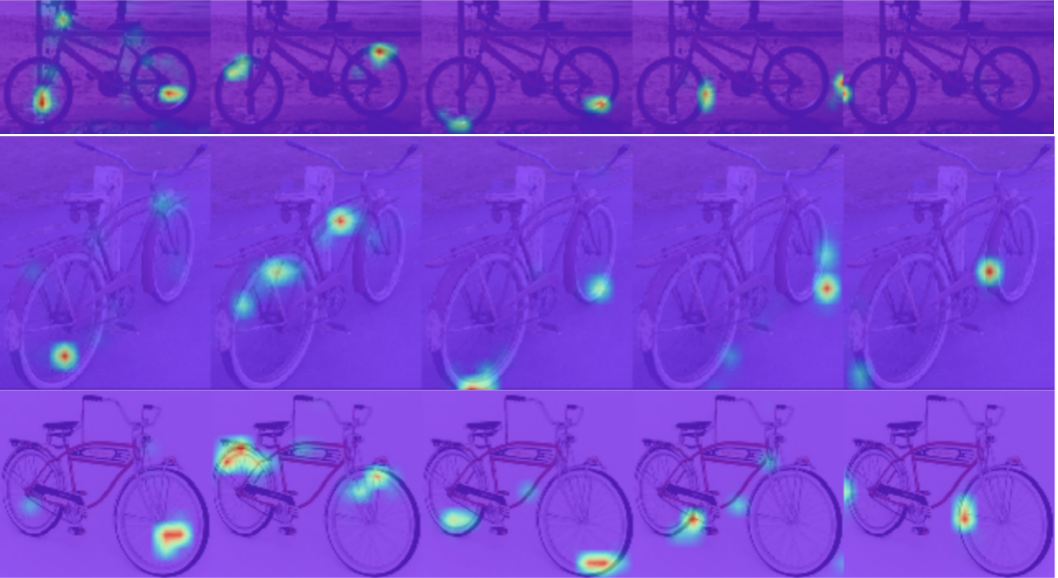}
\caption{The initialized part layer predicts reasonable but not perfect part feature map of bicycle as shown in picture. Each column corresponds to the same output channel and represents one part, from left to right are: {\em wheel center}, {\em upper part of the wheel}, {\em bottom part of the wheel}, {\em front part of the wheel}, {\em back part of the wheel}}
\label{Fig:VC_bicycle}
\end{figure}

As Figure~\ref{fig:test_phase} shown, given an input image $\bm{I}\in\mathbb{R}^{H\times W\times 3}$, our method first uses an ImageNet-pretrained deep convolutional neural network (VGG-16\cite{simonyan2014very} for our implementation) to build a backbone feature map $F\in\mathbb{R}^{h\times w\times c_i}$. Then we add an extra convolutional layer (named part layer) to model and capture parts of the object based on the backbone feature map.
The part layer is initialized by weights generated from clustering of backbone feature vectors~\cite{kortylewski2020combining}.
Then the semantic part information can be extracted in the output feature $\bm{O}\in\mathbb{R}^{h\times w\times c_o}$, after normalized by another layer of softmax. The middle part of Figure~\ref{fig:test_phase} illustrates the backbone of our framework, which maps an input image to the output semantic part map ($\bm{O}=\text{PartLayer}\circ\text{Backbone}(\bm{I})$). The inference of this network is very straightforward and computationally inexpensive. As we will show in the Figure~\ref{Fig:VC_bicycle}, without further training, the initialized 
part layer already learns the semantic parts or key-points reasonably well, with the part layer initialized by the clustering of feature vectors. However, this initialization is far from perfect, for example, firing inconsistently on different images, and firing on different part regions. In the next subsections, we will discuss the training process of the backbone network, without ground-truth supervision, to further tune the part layer for a stronger semantic part discovering.

\begin{figure*}[!ht]
\centering
  \includegraphics[width=0.9\linewidth]{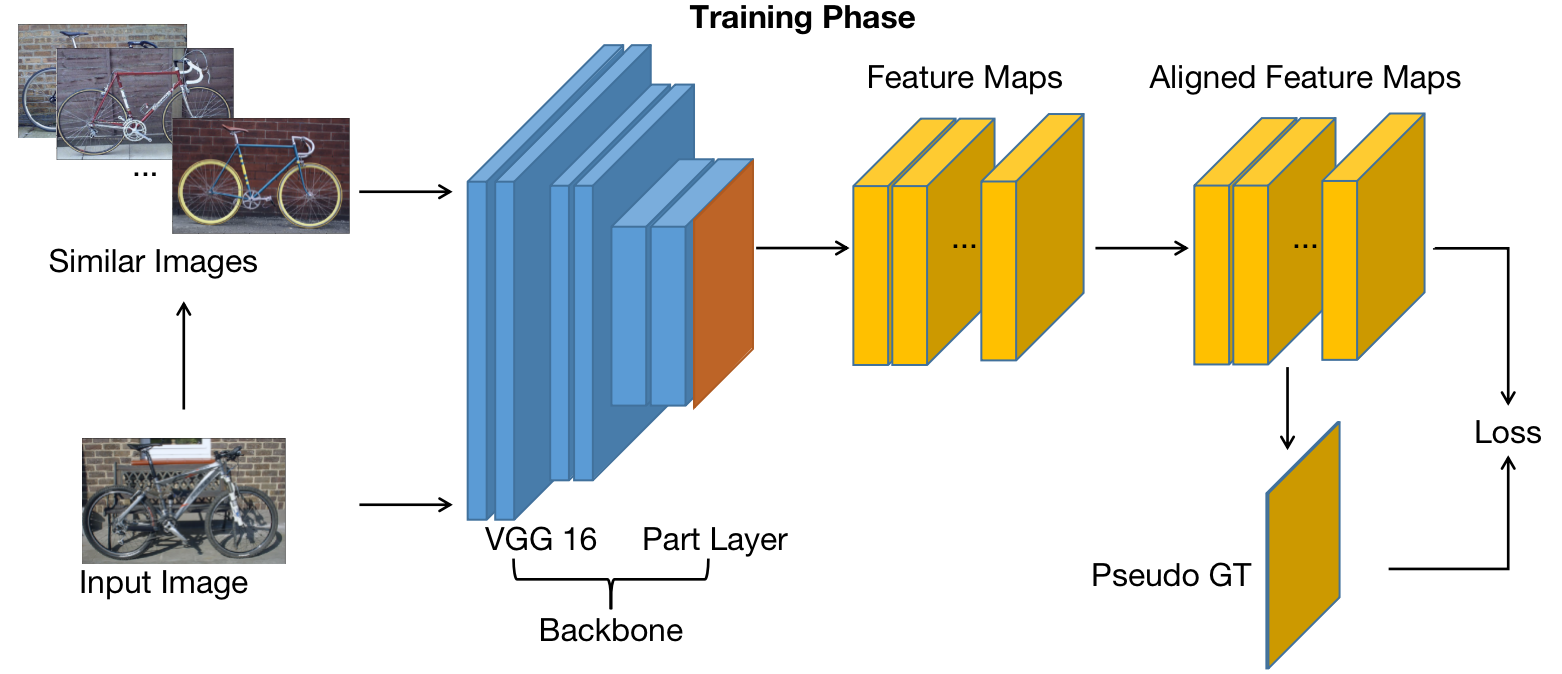}
  \caption{This figure shows the training mechanism, which can be mainly divided into four steps. The first step is to discover similar images by cosine similarity (on the left of figure). The second step is to align feature maps based on the spatial transformation computed via matching (on the upper right). The third step is to generate a fake ground truth (on the bottom right). The last step is to compute the loss between aligned feature maps and pseudo ground truth (on the rightmost of picture). }
  \label{fig:train_phase}
\end{figure*}

\subsection{Training Phase}
The intuition of our work is that similar images can be aligned and the structure of the aligned images should be coherent, thus part detectors should fire coherently on them. To be specific, given a training image $\bm{I}_i\in\mathbb{R}^{H\times W\times 3}$ without knowing its ground-truth parts/key-points, we first want to find a small pool of other training images: $\text{\textbf{SimPool}}(\bm{I}_i)=\{\bm{I}_{i_1}, \bm{I}_{i_2},\cdots,\bm{I}_{i_k}\}$, which are visually and semantically similar to $\bm{I}_i$. By similar, we mean they should not have largely different poses, sub-types, illuminations, \etc. Then the backbone feature maps of all those images are extracted to remove the low-level details and keep the semantics, noted as $\bm{O}_i$ and $\{\bm{O}_{i_1}, \bm{O}_{i_2},\cdots,\bm{O}_{i_k}\}$. Before directly encouraging coherence between these feature maps, we apply some small spatial transformations: $\bm{O}'_{i_j}=\text{\textbf{SpatialTrans}}(\bm{O}_{i_j})$, in order to adjust for the small spatial differences between the training image and the images between the similar pool, and align the feature maps of the similar images to the canvas of $\bm{O}_i$.

Given the aforementioned feature maps $\bm{O}_i$ and $\{\bm{O}'_{i_1}, \bm{O}'_{i_2},\cdots,\bm{O}'_{i_k}\}$, supervision for this training image emerges from encouraging coherence between them, \ie, $$\text{\textbf{Loss}}(\bm{I}_i)=\text{\textbf{Coherence}}(\bm{O}_i, \bm{O}'_{i_1}, \bm{O}'_{i_2},\cdots,\bm{O}'_{i_k})$$
The intuition is that if $\bm{O}_i$ or $\bm{O}'_{ij}$ is largely different at a specific location of a feature channel compared with the other features from the similar image pool, we believe it should be a false positive at that location if the channel is considered semantically meaningful. In the case the feature channel is not semantically meaningful (\eg the initialized part layer is not tight or corresponds to background), the coherence can encourage the feature channel learning something appearing consistently on those images.

Figure~\ref{fig:train_phase} shows the training pipeline of our model, and we will discuss each component in details in the following subsections.

\subsubsection{Similar image finding}
\label{Method:Training:SimilarImage}
\begin{figure}\includegraphics[width=\linewidth]{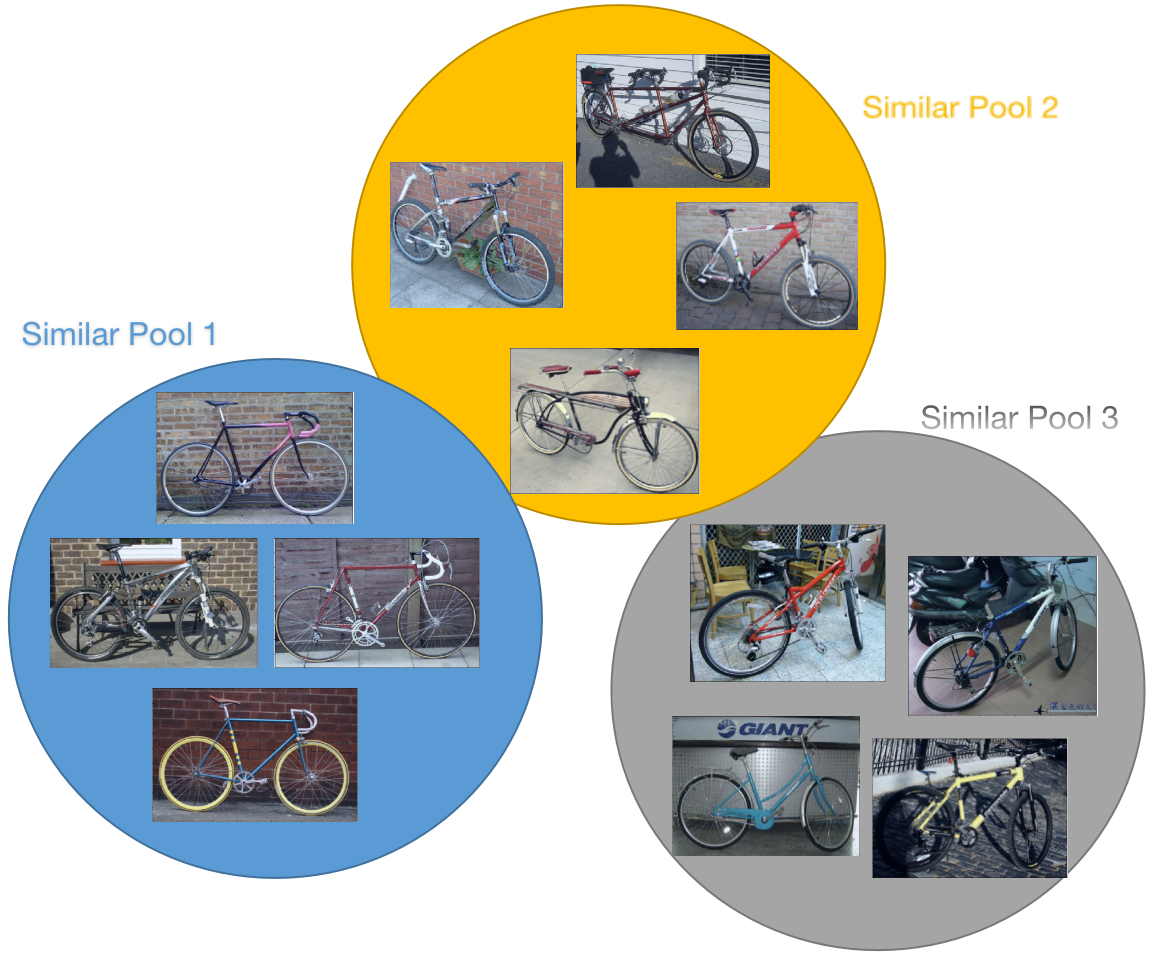}
\caption{Examples of similar image pool for bicycle images of the VehiclePart dataset. The similar images always have similar poses and shapes.}
\label{Fig:similar_pool}
\end{figure}

For each training iteration, the first step is to discover a small pool of images who are similar to the training image, from the full training set. To determine the similarity between two images $\bm{I}_a$ and $\bm{I}_b$, instead of comparing the RGB pixels, we compare the feature maps generating from the initialized model, \ie, $\bm{O}_a$ and $\bm{O}_b$. To be specific, we compare the Jensen-Shannon divergence between the resized feature maps by $s_{ab}=\text{\textbf{Sim}}(\bm{I}_a,\bm{I}_b)=\bm{D}_{JS}(\bm{O}_a,\bm{O}_b)$. The pair-wise similarity scores can be pre-computed for the initialized model, and can also be updated every a few epochs during the training. At the beginning of each training iteration, we first find the small pool of similar images by retrieving the top-k images from the training set (or a subset of the training set in case it is too large), thus: $$\{\bm{I}_{i_1},\cdots,\bm{I}_{i_k}\}=\text{\textbf{SimPool}}(\bm{I}_i)=\{\bm{I}_j: j\in\text{ArgTopK}_j(s_{ij})\}$$
Some examples of the similar image pools are visualized in Figure~\ref{Fig:similar_pool}. This process is very fast since the similarity scores are pre-computed and the similar image pools can also be pre-fetched if necessary.

\subsubsection{Feature map aligning}
\label{Method:Training:Align}
In order to encourage coherence between the training image and the similar images, we want to first align them on the feature map level, to adjust for small variances (\eg small pose/structure/aspect ratio differences). To enhance weak part responses, we first normalize each feature map with its maximum value in the spatial dimensions. Given the training image feature map $\bm{O}_i$, we aim at aligning every image feature map $\bm{O}_{ij}$ extracted from the similar image pool to the canvas of $\bm{O}_i$, via RANSAC. Note that $\bm{O}_i$ and $\bm{O}_{ij}$ may have different shapes (\ie, $h_1\times w_1$ and $h_2\times w_2$).

To estimate spatial transformation and carry out the alignment, we first compute the cosine similarity of feature vectors between the aforementioned feature maps, \ie, $\bm{c}(pq;rs)=\text{\textbf{Cos}}(\bm{O}_i(pq), \bm{O}_{ij}(rs))$, where $p\in[1,h_1],q\in[1,w_1],r\in[1,h_2],s\in[1,w_2]$. A threshold is used to find those matched pairs, \ie $M=\{pqrs: \bm{c}(pq;rs)>th\}$. We randomly select a subset from $M$, and estimate the parameters for a spatial transformation (\eg affine transformation) which maps a coordinate in the similar image feature map to the original training image feature map ($\hat{pq}=T(rs,\theta)$). Number of in-liner counts are computed by how many matched in $M$ satisfy the estimated spatial transformation, and the best transformation with the largest set of in-liner matches are used as the final spatial transform for the corresponding image pair of the training image and the similar image, and the aligned feature map can be obtained by transferring the coordinates on $\bm{O}_{ij}$ to the canvas of $\bm{O}_i$. The aligned feature map of the similar image is: $$\bm{O}'_{ij}=\text{\textbf{SpatialTrans}}(\bm{O}_{ij})=\text{Warp}(\bm{O}_{ij}, T(\cdot,\hat\theta))$$

\subsubsection{Pseudo ground-truth generation}
\label{Method:Training:PseudoGT}
The supervision of our training comes from the coherence between the training image feature map $\bm{O}_i$ and the aligned similar image feature maps $\{\bm{O}'_{ij},j\in[1,k]\}$. In order to encourage the coherence, we first generate a pseudo ground-truth for each output channel from the similar image pool, and then use that to supervise the feature maps as a per-grid segmentation task. The details of the pseudo ground-truth generation is illustrated in Algorithm~\ref{alg:pseudo_GT}. It generates a pseudo ground-truth map of shape $h\times w\times 1$, by first computing the average feature map of $\{\bm{O}'_{ij},j\in[1,k]\}$ and finding the maximum values iteratively in the averaged tensor. Note that for each channel, we allow at most 3 spatial positions corresponding to that, to encourage the discriminativeness. The generated pseudo ground-truth map indicates the coherent firing pattern of the image pool with images similar to the training image.

\begin{algorithm}
    \caption{generate pseudo GT}
    \label{alg:pseudo_GT}
    \begin{algorithmic}[1] % display line number
        \Require $\{\bm{O}'_{ij},j\in[1,k]\}$: the group of aligned feature maps, with shape of $h\times w\times c$ for each of them.
        \Function {generate}{$\{\bm{O}'_{ij},j\in[1,k]\}$}
            \State $\bar{O} \gets \text{Mean}(\{\bm{O}'_{ij},j\in[1,k]\})$
            % \State $\bar{\bm{O}}^i \gets \bar{O}$
            \State $h,w,c \gets \bar{O}.shape$
            \State $\bm{GT}_{pseudo} \gets -ones(h,w)$
            \State $cnt \gets zeros(c)$
            \For{$i = 0 \to h*w$}
                \State $h_i,w_i,c_i = \bar{O}.argmax()$
                \State $\bm{GT}_{pseudo}[h_i,w_i] \gets c_i$
                \State $\bar{O}[h_i,w_i,:] \gets -1$
                \State $cnt[c_i] \gets cnt[c_i] + 1$
                \If{$cnt[c_i] == 3$ and $c_i < c - 1$}
                    \State $\bar{O}[:,:,c_i] \gets -1$
                \EndIf
            \EndFor
            \State \Return{$\bm{GT}_{pseudo}$}
        \EndFunction
    \end{algorithmic}
\end{algorithm}

\subsubsection{Loss function and supervision}
\label{Method:Training:Loss}
With our intuition, the detection training can be supervised by the pseudo ground-truth on the output feature maps.
To encourage the aligned feature maps from different images to be coherent, we simply request all of them to be similar to the pseudo ground-truth. To this end, we adopt a per-pixel classification on the averaged aligned feature maps generated from $\bar{\bm{O}}=\text{Average}(\{\bm{O}_i, \bm{O}'_{i1}, \cdots, \bm{O}'_{ik}\})$. The loss is computed with log likelihood loss $\bm{L}_{nll}$, \ie:
$$
\bm{L} = \bm{L}_{nll}(\log(\bm{\overline{O}}^i), \bm{GT}_{pseudo})
$$

\subsection{Testing Phase}
The test phase is very simple and fast. After the images going through the backbone and normalization, we have feature maps. Then, we generate anchors from feature maps and then select some of them by Non-Maximum Suppression as our prediction. About the evaluation and quantitative results, we will talk about it in Section~\ref{Experiments:Quantitative}.

\section{Experiments}
\label{Experiments}
In this section, we conduct comprehensive experiments on several datasets with landmarks/keypoints for multiple domains such as vehicles, birds and human faces. Our model outperforms several baselines and state-of-the-art alternatives. 
We carry out several ablation experiments that demonstrate the effect of the individual components in our model.
Finally, we analyze qualitative examples which demonstrate that our method is able to discover semantically meaningful part representations in an unsupervised manner.

\subsection{Implementation Details}
We implement our method with PyTorch \cite{paszke2019pytorch}, and we train our models with a single nVIDIA GPU. We use a VGG16 network~\cite{simonyan2014very} as our backbone, and initialize it with ImageNet~\cite{DBLP:journals/ijcv/RussakovskyDSKS15} pre-trained weights. A part learning layer is added after the VGG16 to discover and detect object parts. This layer is initialized with a part dictionary obtained by running an unsupervised clustering algorithm on VGG16 feature vectors on the training dataset~\cite{kortylewski2020combining}. From the illustration in Figure~\ref{Fig:VC_bicycle}, we can observe that these clusters correspond reasonably well to object parts with some errors, which indicates the quality of this strong baseline. As a classifier, we apply a softmax layer on top of the channel-normalized feature vectors.

\subsection{Datasets}

\textbf{VehiclePart.} The VehiclePart~\cite{wang2017detecting} dataset is an extension of Pascal 3D+~\cite{PASCAL} and contains images of vehicles including aeroplanes, bicycles, buses, cars, motorbikes and trains. Each category contains about 300 to 2000 training images and similar amounts of testing images. In addition to the landmark labels provided in Pascal 3D+, the VehiclePart dataset additionally defines more than 10 semantic parts for each vehicle type.

\textbf{CUB} The CUB~\cite{CUB} dataset contains 10,000 bird images with 200 categories of birds with various bird poses and from different camera views. The training set holds 5,994 images while the test set holds 5,794 images. Each image has 15 part landmarks.

\textbf{CelebA.} The CelebA~\cite{celeba} dataset contains about 200k human face images. Following the settings used in~\cite{zhang2018unsupervised}, we use the CelebA train set excluded MAFL~\cite{MAFL} (subset of CelebA) test set as our train set (161962 images in total),

and use the MAFL test set (1000 images in total) as our test set. The dataset provides two different settings, one is unaligned images with annotated bounding box around the face. The other setting involvescroped and aligned faces. We apply our method to both setups. Each image is annotated with 5 facial landmarks.

\textbf{MAFL.} MAFL~\cite{MAFL} is a subset of CelebA. The only difference to CelebA is the size of the training set (19k in MAFL, 162k in CelebA). 

\subsection{Training/Testing Details}
During both training and testing, the short side of the input image is resized to 224 pixels so that all objects are under roughly the same scales. The VGG16 backbone is fixed and only the part discovery layer is trained. The part discovery layer generates 513 channels, including 512 part clusters initialized from cluster centers as discussed in~\cite{kortylewski2020combining} and one background channel.
For the training phase, we use the Adagrad optimizer with a learning rate of $5\times10^{-3}$ and a learning decay of 0.1. The top-k used for building similar image pools is 15, inside a subset of training set with 2000 images. This reduces the memory used for storing the pair-wise similarity scores as discussed in Subsection~\ref{Method:Training:SimilarImage}. The cosine similarity threshold for feature map alignment is 0.6. We use an affine transformation to warp feature maps between similar images.
For the testing phase, the IOU threshold of the non-maximum suppression is 0.3 during the post-processing.

\subsection{Quantitative Results}
\label{Experiments:Quantitative}

In this section, we discuss quantitative results and comparisons, and illustrate that our novel idea is promising and our innovative designed method is more effective than other baselines.

First, following the designs in the VehiclePart dataset, we use mean average precision (mAP) with an IOU threshold of 0.5 as our evaluation on the part bounding boxes provided in the VehiclePart. Moreover, we use mAP with an L2 distance threshold of 0.1 as the performance metric for both the CUB and CelebA dataset. In this evaluation, our method will be compared with clustering based part discovery~\cite{kortylewski2020combining} and InterpretableCNN~\cite{zhang2018interpretable}. The clustering based part discovery performs unsupervised clustering on the feature vectors from a backbone CNN, and uses these cluster centers as part detectors. Regarding the InterpretableCNN, it adds a regularizer on a traditional CNN, by assigning spatial masks to each channel of the CNN and using the corresponding mask to supervise and regularize the CNN. Thus those channels can better correspond to parts. We call them Clustering and MaskRegularized in the remainder of this paper.

In addition to the aforementioned baselines, we also compare our results on the CelebA and CUB dataset with recent unsupervised methods~\cite{hung2019scops, huang2020interpretable, thewlis2017unsupervised_59, zhang2018unsupervised} and other supervised methods~\cite{zhang2014coarse, zhang2015learning, sun2013deep, zhang2014facial}. In these comparisons, we follow their evaluation protocol and use L2 distance normalized by either the pupil distance (for CelebA) or the object bounding box size (for CUB). In the experiments, we follow the common mechanism of using a linear regressor trained on part of the training set to map the unsupervised predictions to the manually labeled landmarks.

\begin{table}[!tb]
  \centering
  \scriptsize
  \setlength{\tabcolsep}{0.00cm}
  \begin{tabular}{C{1.5cm}|C{0.9cm}C{0.9cm}C{0.9cm}C{0.9cm}C{0.9cm}C{0.9cm}C{0.9cm}}
  \shline
  mAP & Aero & Bicycle & Bus & Car & Mbike & Train & Average \\ \shline
  Clustering  & $16.7$ & $54.6$ & $28.7$ & $32.2$ & $32.2$ & $17.6$ & $29.7$ \\ \hline
  MaskReg & $\mathbf{18.7}$ & $56.9$ & $37.4$ & $29.4$ & $37.9$ & $\mathbf{21.0}$ & $33.6$\\ 
  \shline
  Ours & $\mathbf{18.7}$ & $\mathbf{66.7}$ & $\mathbf{46.6}$ & $\mathbf{35.5}$ & $\mathbf{39.6}$ & 19.4 & $\mathbf{37.8}$ \\
  \hline
\end{tabular}
\caption{Results of unsupervised part detection methods on VehiclePart, including {\em aeroplane}, {\em bicycle}, {\em bus}, {\em bus}, {\em motorbike}, and {\em train}. Our model can predict bounding boxes more accurately than Clustering and MaskRegularized methods (see details in the second paragraph of Section~\ref{Experiments:Quantitative}).}
\label{tab:PASCAL}
\end{table}

\subsubsection{VehiclePart}
In the Table~\ref{tab:PASCAL}, we show the mean AP with IOU threshold of 0.5 on the VehiclePart dataset. Comparing with the mAP of $29.7$ for the Clustering method and the mAP of $33.6$ for the MaskRegularized method, our average performance of the six categories is $37.8$ mAP, which achieves an improvement of at least 12.5\%. The results illustrate that our method is promising compared with other unsupervised methods. Different from the Clustering method, we take both local appearance of the part in one image and inter-image part alignment into consideration when learning the part dictionary (\ie the part layer weights). The MaskRegularized method mainly uses single peak feature maps to regularize the CNN channels, and the supervision is not very strong. For example, during the training, a bad channel feature map may lead to a bad mask. In addition, it only allows for a single peak per channel for each training example, while our method allows for multiple peaks (up to three) in one feature map, which can help for discovering parts such as the wheels of a car.

\begin{table}
\centering{
\small
\setlength{\tabcolsep}{0.08cm}
\begin{tabular}{C{2cm}|C{2cm}C{2cm}}
\shline
 {mAP} &  Unalign & Align \\
\hline
 Clustering & $71.8$ & $92.6$ \\
\shline
 Ours & $\mathbf{90.9}$ & $\mathbf{97.9}$ \\
\hline
\end{tabular}}
\caption{Compared with our baseline on CelebA aligned and unaligned set (both 200k images, including 5 parts), our result is 19.1 or 5.3 points ahead.}
\label{Tab:CelebA}
\vspace{-0.4cm}
\end{table}

\begin{table}
\centering{
\small
\setlength{\tabcolsep}{0.08cm}
\begin{tabular}{C{2cm}|C{2cm}C{2cm}}
\shline
 {Error} & Unalign & Align \\
\shline
 Clustering\cite{wang2017visual} & $13.8$ & $6.0$ \\
\hline
 \cite{hung2019scops} & $15.0$ & $-$ \\
\hline
 \cite{huang2020interpretable} & $8.4$ & $-$ \\
\hline
 \cite{thewlis2017unsupervised_59} & $-$ & $5.33$ \\
\hline
 \cite{zhang2018unsupervised} & $-$ & $\mathbf{3.15}$ \\
\shline
 Ours & $\mathbf{6.3}$ & $5.0$ \\
\hline
\end{tabular}}
\caption{{Results of applying unsupervised landmarks detection methods on CelebA aligned and unaligned set (both 200k images, including 5 parts). Our result achieves state-of-the-art on unaligned set and is close to \cite{zhang2018unsupervised} on align set.}}
\label{Tab:CelebA_LR}
\vspace{-0.4cm}
\end{table}

\begin{table}
\centering{
\small
\setlength{\tabcolsep}{0.08cm}
\begin{tabular}{C{3cm}|C{3cm}|C{2cm}}
\shline
\multicolumn{2}{C{6cm}|}{Error} &  MAFL align \\
\shline
\multirow{4}{*}{Fully supervised} & CFAN\cite{zhang2014coarse} & $15.84$ \\
 {} & TCDCN\cite{zhang2015learning} & $7.59$ \\
 {} & Cascaded CNN\cite{sun2013deep} & $9.73$ \\
 {} & MTCNN\cite{zhang2014facial} & $\mathbf{5.39}$ \\
\hline
\multirow{4}{*}{Unsupervised} & \cite{thewlis2017unsupervised_59} & $6.67$ \\
 {} & \cite{thewlis2017unsupervised_58} & $5.8$ \\
 {} & \cite{zhang2018unsupervised} & $\mathbf{3.15}$ \\
 {} & Ours & $4.9$ \\
\hline
\end{tabular}}
\caption{\textcolor{black}{Compared with both unsupervised and supervised part detection methods on MAFL aligned set, our result surpasses all supervised methods and only worse than \cite{zhang2018unsupervised} of unsupervised methods.}}
\label{Tab:MAFL}
\vspace{-0.4cm}
\end{table}

\subsubsection{CelebA human face dataset}
To further verify the effectiveness of our method and its applicability to other datasets, we test our method on CelebA, which has a training set of 80 times larger compared with the VehiclePart dataset. In Table~\ref{Tab:CelebA}, we report the mean AP based on the L2 distance threshold. Compared with the Clustering method, our method improves $19.1\%$ on the unaligned setting and $5.3\%$ on the aligned setting. Hence, we observe that our method a similar effect as on the VehiclePart dataset, namley that our method improves the part layer weights.

In addition, we also compare our method with the state-of-the-art unsupervised methods. In the Table~\ref{Tab:CelebA_LR}, we evaluate results by L2 distance normalized by pupil distance after linear regression and obtain the error of $6.3\%$ for unaligned and $5.0\%$ for aligned images. Notably, we achieve the state-of-the-art performance in the more challenging unaligned setting. 
\cite{zhang2018unsupervised} is slightly better ($1.8\%$ of pupil distance) compared with our method in the aligned setting.

We surpass all supervised methods on the MAFL dataset in the aligned setting (with an error of $4.9\%$), while being slightly worse than \cite{zhang2018unsupervised} when comparing with unsupervised methods as shown in Figure~\ref{Tab:MAFL}.

\begin{table}
\centering{
\small
\setlength{\tabcolsep}{0.08cm}
\begin{tabular}{C{2cm}|C{2cm}}
\shline
 mAP & Bird \\
\shline
 Clustering & $26.9$ \\
\shline
 Ours & $\mathbf{28.2}$ \\
\hline
\end{tabular}}
\caption{\textcolor{black}{Compared with baseline on CUB 10k images of 200 categories including 15 parts, our result is 1.3 points better.}}
\label{Tab:CUB}
\vspace{-0.4cm}
\end{table}

\begin{table}
\centering{
\small
\setlength{\tabcolsep}{0.08cm}
\begin{tabular}{C{2cm}|C{2cm}}
\shline
 {Error} & Bird \\
\hline
 Clustering & $13.9$ \\
\hline
 \cite{huang2020interpretable} & $11.5$ \\
\shline
 Ours & $\mathbf{11.1}$ \\
\hline
\end{tabular}}
\caption{\textcolor{black}{Results of applying unsupervised landmarks detection methods on CUB. Our result reaches the best results among them.}}
\label{Tab:CUB_LR}
\vspace{-0.4cm}
\end{table}

\begin{table*}[htb]
\centering{
\small
\setlength{\tabcolsep}{0.08cm}
\begin{tabular}{C{2.8cm}|C{2.8cm}|C{1.5cm}C{1.5cm}C{1.5cm}C{1.5cm}C{1.5cm}C{1.5cm}C{1.5cm}}
\shline
mAP & {Configuration} & {Aeroplane} & {Bicycle} & {Bus} & {Car} & {Motorbike} & {Train} & Average \\
\shline
\multirow{4}{*}{Top-k} & 5 & $16.4$ & $64.6$ & $41.9$ & $32.6$ & $\mathbf{39.9}$ & $14.4$ & $35.0$ \\
 {} & 10 & $16$ & $66.2$ & $46.5$ & $\mathbf{37.3}$ & $38.8$ & $\mathbf{20.9}$ & $37.6$ \\
 {} & 15 & $\mathbf{18.7}$ & $\mathbf{66.7}$ & $46.6$ & $35.5$ & $39.6$ & $19.4$ & $\mathbf{37.8}$ \\
 {} & 20 & $18.5$ & $64.1$ & $\mathbf{48.3}$ & $36.5$ & $36.7$ & $19.2$ & $37.2$ \\
\hline
\multirow{3}{*}{Spatial transform} & affine & $16$ & $66.2$ & $\mathbf{46.5}$ & $\mathbf{37.3}$ & $38.8$ & $\mathbf{20.9}$ & $\mathbf{37.6}$ \\
 {} & translation & $\mathbf{18.7}$ & $\mathbf{68.6}$ & $44.5$ & $36.5$ & $\mathbf{40.3}$ & $16.6$ & $37.5$ \\
 {} & homography & $16.9$ & $65.3$ & $45.6$ & $35.7$ & $40.2$ & $13.1$ & $36.1$ \\
 {} & without transform & $15.3$ & $64.4$ & $42.2$ & $33.3$ & $36.4$ & $16.5$ & $34.7$ \\
\hline
\multirow{3}{*}{Suppress neighbour} & 0 & $\mathbf{16}$ & $66.2$ & $46.5$ & $\mathbf{37.3}$ & $38.8$ & $\mathbf{20.9}$ & $37.6$ \\
 {} & 1 & $15.6$ & $69.2$ & $\mathbf{47.3}$ & $34.3$ & $\mathbf{41.9}$ & $17.1$ & $37.6$ \\
 {} & 2 & $15.9$ & $\mathbf{73.4}$ & $45.7$ & $\mathbf{37.3}$ & $41.6$ & $15.3$ & $\mathbf{38.2}$ \\
\hline
\end{tabular}}
\caption{\textcolor{black}{This figure shows three ablation studies about different unit in our pipeline. The detailed analysis is discuss in Section~\ref{Ablation Studies}}}
\label{Tab:Ablation}
\vspace{-0.4cm}
\end{table*}

\subsubsection{CUB dataset}
\begin{figure}\includegraphics[width=\linewidth]{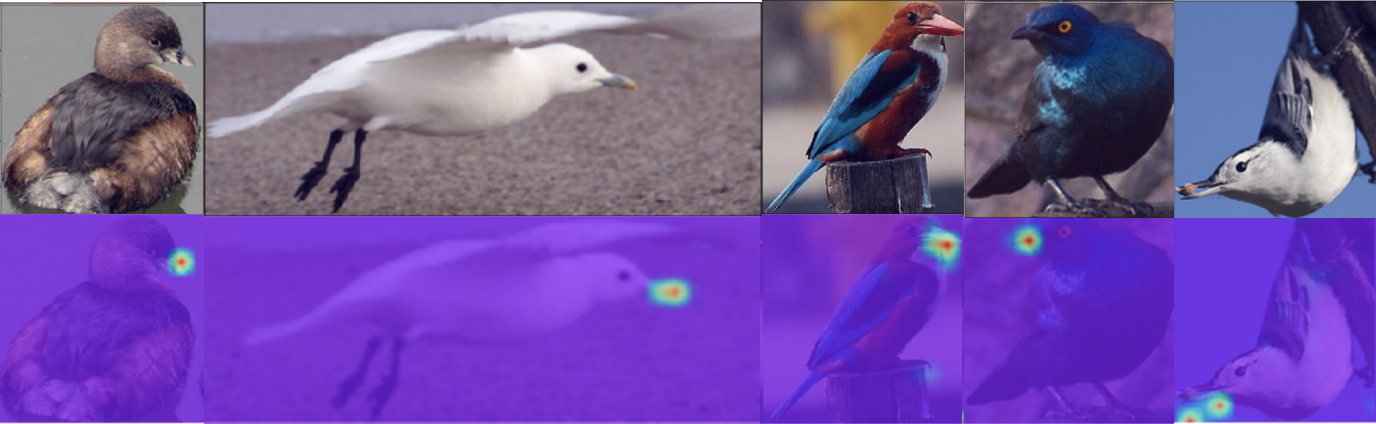}
\caption{CUB dataset contains birds with different shapes and colorful feathers, various bird pose, multiple camera viewpoints. We take the {\em beak} part as an example. Our method can discover parts clearly and robustly.}
\label{Fig:bird}
\end{figure}

The objects of VehiclePart and CelebA datasets are considered rigid objects (\ie, vehicles and faces), so we would like to verify our method's ability to discovery non-rigid objects parts/landmarks, such as deformable birds. The CUB dataset is challenging because it consists of 200 bird classes in various poses (see examples in Figure~\ref{Fig:bird}), \eg, standing, swimming, or flying, as well as the different camera viewpoints. Our performance is $28.2\%$ for mAP as shown in the Table~\ref{Tab:CUB} while the Clustering method achieves $26.9\%$. Notably, the~\cite{huang2020interpretable} method trains with the image-level label while our method is free of using these extra labels. Despite using less supervision, we still achieve state-of-the-art performance as shown in Table~\ref{Tab:CUB_LR}.

\begin{figure}\includegraphics[width=\linewidth]{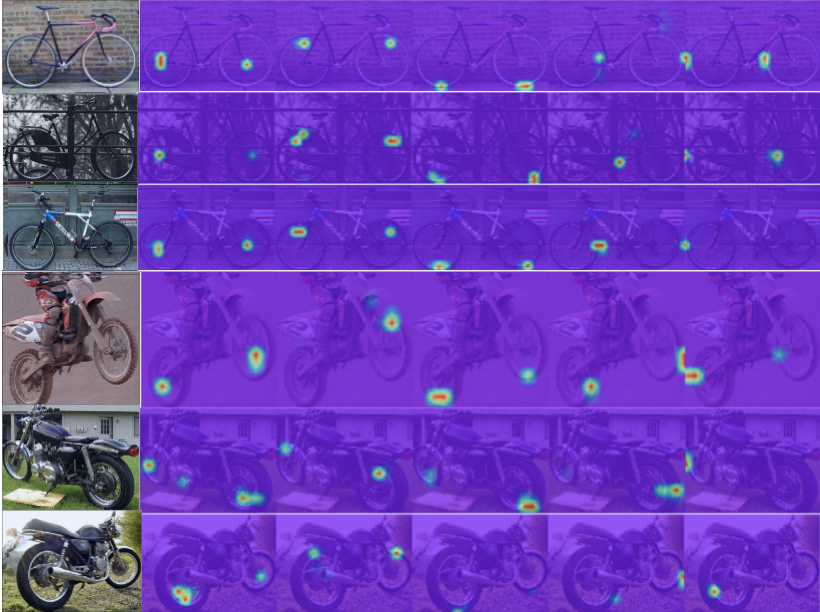}
\caption{Visualization of the bicycle (row 1-3) and the motorbike (row 4-6). We show the corresponding feature map for the first five parts: {\em wheel center}, {\em upper part of the wheel}, {\em bottom part of the wheel}, {\em front part of the wheel}, {\em back part of the wheel}.}
\label{Fig:bicycle_motorbike}
\end{figure}

\begin{figure}\includegraphics[width=\linewidth]{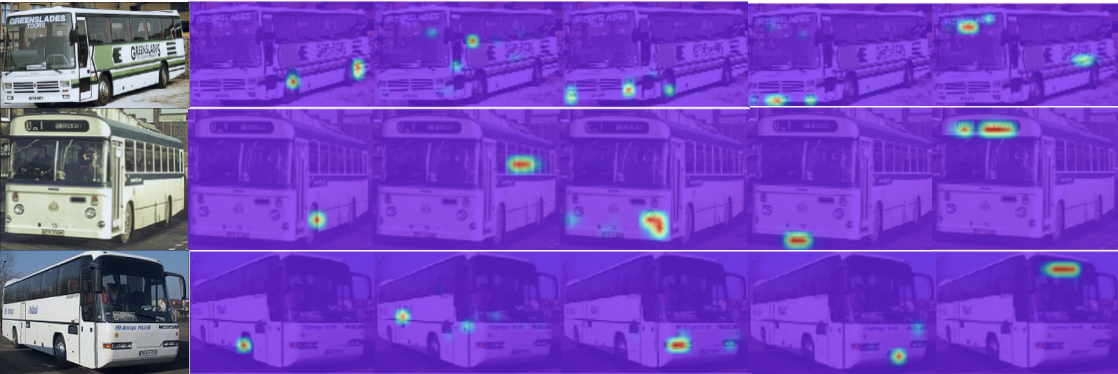}
\caption{Visualization of the bus. We show the corresponding feature map for the first five parts: {\em wheel}, {\em side window and side body}, {\em headlight}, {\em license plate}, {\em display screen}.}
\label{Fig:bus}
\end{figure}

\begin{figure}\includegraphics[width=\linewidth]{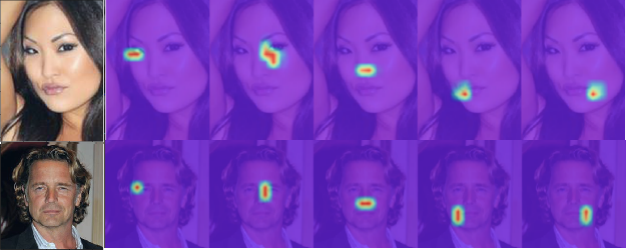}
\caption{Visualization of the human face (row 1 is unaligned and row 2 is aligned). We show all five parts feature map: {\em left eye}, {\em right eye}, {\em nose}, {\em left mouth}, {\em right mouth}}
\label{Fig:MAFL}
\end{figure}

\subsection{Ablation Studies}
\label{Ablation Studies}
To explore the individual contribution of each component in our pipeline, we conduct a detailed ablation study and illustrate the performance in Table~\ref{Tab:Ablation}. To be more specific, we mainly investigate three different components in our method and will discuss them in the following paragraphs.

Firstly, top-k in similar images finding, which is the number of the most similar images compared with the given training image, is important to our training. The performance will drop by more than 2 points when we use a smaller similar image pool with only 5 images. When the top-k is too big (\eg goes up to 20), the performance may drop a little bit, in the case that the training sets for some categories are not large enough to provide a large amount of similar images.

Second one is a transformation in the feature map alignment unit. As the results shown, our method has a good robustness to different transformations. Even using simple translation, we can achieve promising results. On the other hand, without transformation to align feature maps, the performance drops by about 3 points. The conclusion is that our idea about alignment is necessary to provide solid supervision.

We also diagnose the process of  pseudo ground-truth generation. In order to comply with the dataset that one part may have more than one instance (\eg, a car usually has more than one wheel), we allow multiple locations for a single channel when building the pseudo ground-truth. To avoid selecting redundant nearby locations, we suppress the average feature map values for the selected channel around the selected location. The radius to be suppressed are 0, 1 and 2, and we report the performance for these settings in the third block of Table~\ref{Tab:Ablation}. We can see the results are significantly improved for bicycle and motorbike with the suppression. However, other categories may have some regressions. So we choose to use a suppression radius of 0 in our final settings.

\begin{table}
\centering{
\small
\setlength{\tabcolsep}{0.08cm}

\begin{tabular}{C{2cm}|C{2cm}|C{2cm}}
\shline
 {error} & {\em unalign} & {\em align} \\
\hline
 CelebA & $6.1$ & $\mathbf{5.0}$ \\
\hline
 MAFL & $\mathbf{7.1}$ & $4.9$ \\
\hline
\end{tabular}}
\caption{\textcolor{black}{We research about the influence of using different size of training set to our method. The results illustrate that our method is not sensitive to quantity}}
\label{Tab:Ablation_number}
\vspace{-0.4cm}
\end{table}

Finally, to verify our method robust on small training size, we train our model and fit the linear regressor both on MAFL train set and test on MAFL test set. The performances in Table~\ref{Tab:Ablation_number} show that our method is stable when the number of training images decreases.

\subsection{Qualitative Results}
As shown from Figure~\ref{Fig:bicycle_motorbike} to Figure~\ref{Fig:bird}, our method provides feature maps that have strong relation with parts, even those with multiple instances. If we look closely at the Figure~\ref{Fig:bus}, the $5th$ part (last row), display screen located on the upper part of the bus head, can be detected in the right location even without any real display screen there as the first and third bus shown. That prediction proves that our method based on alignment is effective and it can predict part location exactly only based on other images' information. Besides, Figure~\ref{Fig:bird} verifies that our method is robust to object shape, posture, and viewpoints.

\section{Conclusion}
\label{Conclusion}
In this paper, we investigate the problem of unsupervised parts or landmarks discovering, which is a very important topic to understand images better without extra expensive and time-consuming annotations. Based on the intuition that the detected parts should fire in a similar manner on different images, we propose a novel method which first finds and aligns a set of similar images, and then encourages the coherence between the feature maps from those images, to make them to learn tighter and more consistent parts. The proposed method is very simple and efficient during the inference phase, which is simply a feed-forward of a CNN. Extensive experiments have been carried out on several different datasets and domains, and our method is consistently better than our baselines and other alternatives. Further works including unsupervisedly parsing objects or using the parts to improve classification/occlusion could be done based on our work, and we remain them as the future works.

{\small
\bibliographystyle{IEEEtran}
\bibliography{egbib.bib}
}

\end{document}